\documentclass[letterpaper, 10 pt, journal, twoside]{IEEEtran}

\usepackage{times}
\usepackage{dsfont}
\usepackage{amsmath}
\usepackage{amssymb}
\usepackage{multirow}
\usepackage{mathtools}
\usepackage{units}
\usepackage{float}
\usepackage[]{units}
\usepackage{xcolor}
\usepackage{xcolor}
\usepackage{graphicx}

\usepackage[printonlyused]{acronym}

\usepackage{makecell}

\usepackage{hyperref}

\newcommand\revise[1]{{\leavevmode\color{black}{#1}}} 

\newacro{MDP}{Markov Decision Process}
\newacro{CMDP}{Constrained Markov Decision Process}
\newacro{RL}{Reinforcement Learning}
\newacro{PPO}{Proximal Policy Optimization}

\newacro{MPC}{Model Predictive Control}
\newacro{MLP}{Multi Layer Perceptron}
\newacro{RNN}{Recurrent Neural Network}
\newacro{GRU}{Gated Recurrent Unit}
\newacro{CNN}{Convolutional Neural Network}

\newacro{ReLU}{Rectified Linear Unit}
\newacro{GAE}{Generalized Advantage Estimation}
\newacro{P3O}{Penalized Proximal Policy Optimization}

\newacro{IK}{Inverse Kinematics}
\newacro{SDF}{Signed Distance Field}

\newacro{BCE}{Binary Cross-Entropy}
\newacro{RRT}{Rapidly-exploring Random Tree}

\newacro{ESDF}{Euclidean Signed Distance Function}

\newacro{APF}{Artificial Potential Field}
\newacro{TO}{Trajectory Optimization}
\newacro{MPPI}{Model Predictive Path Integral}

\newacro{EE}{End-effector}

\begin{document}

\title{
Learning Fast, Tool-aware Collision Avoidance \\ for Collaborative Robots
}

\author{Joonho Lee$^{*}$, Yunho Kim, Seokjoon Kim, Quan Nguyen, Youngjin Heo
\thanks{This research has been funded by the Industrial Technology Innovation Program (P0028404) of the Ministry of Industry, Trade and Energy of Korea.}
\thanks{All authors are with Neuromeka Co., Ltd.}
\thanks{*\textit{Senior and Corresponding Author (joonho.lee@neuromeka.com)}}
}


\maketitle

\begin{abstract}
\revise{
Ensuring safe and efficient operation of collaborative robots in human environments is challenging, especially in dynamic settings where both obstacle motion and tasks change over time. Current robot controllers typically assume full visibility and fixed tools, which can lead to collisions or overly conservative behavior. In our work, we introduce a tool‑aware collision avoidance system that adjusts in real time to different tool sizes and modes of tool-environment interaction. Using a learned perception model, our system filters out robot and tool components from the point cloud, reasons about occluded area, and predicts collision under partial observability. We then use a control policy trained via constrained reinforcement learning to produce smooth avoidance maneuvers in under 10 milliseconds. In simulated and real‐world tests, our approach outperforms traditional approaches (APF, MPPI) in dynamic environments, while maintaining sub-millimeter accuracy. 
Moreover, our system operates with approximately \unit[60]{\%} lower computational cost compared to a state-of-the-art GPU-based planner. Our approach provides modular, efficient, and effective collision avoidance for robots operating in dynamic environments. We integrate our method into a collaborative robot application and demonstrate its practical use for safe and responsive operation.
}
\end{abstract}

\section{INTRODUCTION}

\IEEEPARstart{T}{he} use of robotic manipulators in contact-rich, real-world applications is rapidly expanding, driven by recent advances in control strategies and hardware~\cite{zhang2024learning, fu2024mobile}.
 These developments have broadened robot deployment into a wide range of settings, from industrial sites to in-home service applications.
 As robots become more integrated into everyday environments, ensuring safety is becoming critical—especially for collaborative and service robots that operate alongside humans and cannot be isolated behind protective barriers like industrial machines.

Most industrial systems rely on collision detection~\cite{haddadin2017robot, heo2019collision} and compliance control~\cite{ko2024compensation, han2019collision}, responding only after a collision has occurred. 
While widely adopted~\cite{proia2021control}, these approaches fall short in dynamic environments populated by humans.
A common preventive measure is to define protective zones around the robot during the pre-impact phase~\cite{organizacion2016iso}. Sensors like LiDAR or safety-rated scanners create "red" zones to trigger safety stops and "yellow" zones to reduce speed. However, these methods do not change trajectories according to obstacles and are limited because robots with high kinetic energy require large safety zones.
Recent works have introduced more advanced perception systems capable of adapting trajectories in real time by measuring distances and maintaining safe separation~\cite{magrini2017human, casalino2019optimal, de2017multimodal, sloth2018computation, sundaralingam2023curobo, faroni2022safety}.

Existing approaches still face limitations in real-world applications. They often rely on fixed geometric parameters and cannot adapt to dynamic environments. They also overlook \emph{controlled contact}, reducing effectiveness in applications like assembly and pick-and-place. Furthermore, their lack of robustness to occlusions and reliance on heuristics often increase system complexity.

\begin{figure}[t!]
    \centering
    \includegraphics[width=0.98\columnwidth]{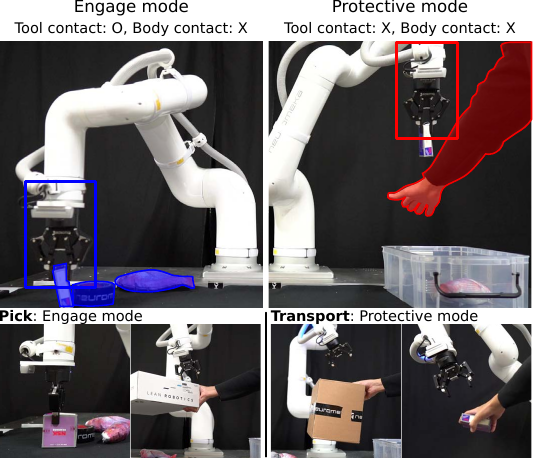}
\caption{
\revise{\textbf{Pick-and-place example.} Our system manages robot-environment interactions by taking an \emph{interaction mode} input that determines contact permissions.
(Left) The tool area is allowed to make contact with the environment during active use.
(Right) At other times, contact is prohibited for both the tool and the robot.}
}
\label{fig:intro} 
\end{figure}

In this work, we propose an adaptive collision avoidance system that provides two command modes for manipulator-environment interaction (see Fig.~\ref{fig:intro}). 
\emph{Engage mode} allows the tool (e.g., gripper) to make contact during active tasks, while \emph{Protective mode} prevents all contact during transport phases. This tool-aware design enhances adaptability and enables safe and efficient operation in dynamic contact-rich settings.

\subsection{System Requirements}
We define three key requirements for an effective collision avoidance system in real-world applications:
\begin{itemize} 

\item \textbf{Flexibility to Tool Settings:}
Robots use a wide range of \ac{EE}s, so the system should adapt to varying tool shapes and operational modes in real time.

\item \textbf{Precision:}
The system should achieve high precision in nominal operations, ensuring consistent performance.

\item \textbf{Responsiveness and Reliability:}

The system must operate with minimal delays to avoid unsafe conditions, remain interpretable, and remain stable under challenging conditions such as occlusions.

\end{itemize}

\begin{figure}[t!]
    \centering
    \includegraphics[width=\columnwidth]{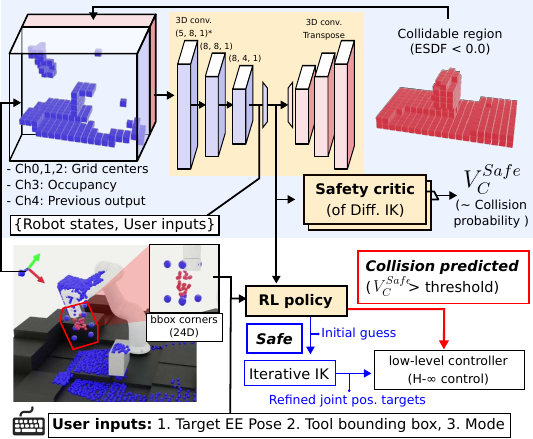} 
\caption{\textbf{System Overview.}
The RL policy generates target joint positions, used as initial guess for IK when safe. 
A 3D CNN encoder processes workspace occupancy and proprioceptive inputs to estimate collision probabilities via a learned safety critic. When the risk is high, RL policy is triggered. All components are trained in simulation.
}
    \label{fig:overview} 
\end{figure}

\revise{
\subsection{System Overview}

Considering the requirements above, we developed a collision avoidance system that combines the robustness of learning-based methods with the precision of a classical controller.
An overview of our system is shown in Fig.~\ref{fig:overview}. The user provides three inputs: the desired \ac{EE} pose, the geometry of the tool attached to the robot and the interaction mode (Engage or Protective mode). 
In nominal conditions, the system solves \ac{IK} using a classical method.
A learned safety critic continuously monitors the environment and evaluates the risk of future collisions.
When the risk exceeds a defined threshold, the system switches to a \ac{RL} policy for reactive motion generation. 

The first part of our system is an encoder-decoder model that processes point cloud data along with its previous occupancy prediction.
The module has two outputs: occupancy grid and safety critic values~\cite{srinivasan2020learning} for both body and tool collision constraints (explained later).

The safety critic value estimates the likelihood of future constraint violations (collisions), based on Srinivasan et al.~\cite{srinivasan2020learning}.
It outputs a scalar value between 0 and 1.
When the safety critic value is below a threshold (0.8 in our implementation), the RL policy output is used as an initial guess for an iterative \ac{IK} method for high-precision.
Conversely, when the critic indicates a high risk, the policy output is directly used for reactive collision avoidance.

The policy uses the latent space of the encoder as its input. This state encapsulates the spatial information necessary for effective collision avoidance~\cite{hoeller2024anymal}. The collision avoidance policy operates at a frequency of \unit[50]{Hz} and generates joint position target residuals.
}


\subsection{Contributions}

Our key advancements are:

\begin{itemize}
\item \textbf{Fast and Adaptive Collision Avoidance:}
\revise{
Our system adapts in real time to user inputs and robustly handles dynamic obstacles. Our learned models adapts their outputs based on tool geometry and modes.
Its low computational cost enables fast, safe execution of contact-rich tasks.
}

\item \textbf{Seamless Integration with High Accuracy:} 
\revise{To balance responsiveness and precision, a fast but less accurate RL policy is activated only under high collision risk.}

\item \textbf{Robust Perception:} 
\revise{Our learned perception model removes the need for explicit robot filtering~\cite{de2017multimodal}, mapping~\cite{oleynikova2017voxblox}, and hand-crafted occlusion heuristics.  This results in a simpler implementation and enables reliable operation in dynamic, partially observable environments.}

\end{itemize}

\revise{We validate our system t}hrough real-world experiments and qualitative evaluations. \revise{Our approach achieves} high precision and reliability while operating an order of magnitude faster than state-of-the-art methods~\cite{sundaralingam2023curobo,liu2022regularized,storm2021}.

\section{RELATED WORK}
We summarize related works in the areas relevant to our research.

\subsection{Learned Scene Representation}
Our approach aligns with recent works that employ neural networks to process raw sensor data for scene representation~\cite{hoeller2024anymal, kulkarni2024reinforcement,miki2024learning,yang2022real,luo2024pie}.
These methods utilize encoder-decoder architectures to process exteroceptive measurements
and use the resulting latent vectors as inputs for policy networks.
Hoeller et al.\cite{hoeller2024anymal} used a 3D \ac{CNN} to handle noisy point clouds, producing a latent representation for navigation and terrain reconstruction. Similarly, Kulkarni et al.\cite{kulkarni2024reinforcement} developed a Deep Collision Encoder (DCE) to extract collision information from noisy depth images for quadrotors using an architecture inspired by variational autoencoders.

These works demonstrate that processing sensor measurements with neural networks to build implicit environmental maps, rather than relying on conventional mapping, leads to highly responsive and robust systems.

\subsection{Collision Avoidance}
Traditional approaches to safe motion planning focus on finding collision-free geometric paths using methods such as \ac{RRT}~\cite{lavalle1998rapidly} and \ac{APF}~\cite{khatib1986real}. Subsequent research has explored \ac{TO} techniques, like CHOMP~\cite{ratliff2009chomp}.

Recent research frames robot navigation as a global optimization problem.
To handle complex environments with multiple possible solutions, sampling-based approaches are employed. For example, Curobo~\cite{sundaralingam2023curobo} combines sampling-based planning and \ac{TO} for robotic manipulators.

These model-based methods generally perform well in fully observable environments where obstacles and dynamics are predictable.
However, they struggle in dynamic, partially observable environments.
\revise{Occlusions and the need for fast replanning make reliable collision avoidance challenging in real-world settings.}

Recently, data-driven approaches, particularly those using \ac{RL}, have emerged as alternatives. These methods allow robots to learn safe maneuvers by quickly responding to dynamic environment~\cite{lee2024learning, thumm2022provably, kulkarni2024reinforcement, kim2024armor}.

\subsection{Safe Reinforcement Learning}

Safe \ac{RL} is a well-established research area, with various approaches differing in their architectural frameworks~\cite{garcia2015comprehensive}.
One widely used approach is the Conditional Value at Risk (CVaR) formulation~\cite{heger1994consideration, geibel2005risk, tang2019worst}.
This method focuses on maximizing the expected return in the lower tail of the return distribution, making it effective for addressing rare but critical events. However, we opted not to adopt this approach due to the complexity involved in implementing distributional critics and tuning risk parameters~\cite{schneider2024learning}, leaving it as a consideration for future work.

In this work, we adopt the \ac{CMDP} framework for its simplicity in implementation and deployment, leveraging its similarity to the widely used PPO algorithm. The \ac{CMDP} framework~\cite{altman2021constrained} offers a structured and efficient approach to incorporating constraints into RL. By defining constraints as cost functions, CMDPs allow the optimization of policies that balance safety and performance~\cite{shen2022penalized, liu2020ipo}. These methods efficiently manage multiple constraints. Kim et al.\cite{kim2024not} and Lee et al.\cite{lee2023evaluation} demonstrated the utility of CMDPs in training safe locomotion policies for quadrupedal robots on rough terrain.



\section{METHOD}
\begin{figure}
    \centering
    \includegraphics[width=\columnwidth]{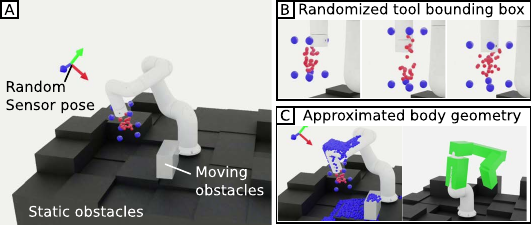} 
\caption{\textbf{Simulation Environment.}
}
    \label{fig:sim} 
\end{figure}

We provide an overview of our system and describe each component in detail.

\subsection{Problem Definition}

Figure~\ref{fig:sim}A illustrates the target environment. We focus on 2.5D environments, where obstacles are primarily oriented in the vertical ($z$) direction with varying heights.
We consider random static and dynamic obstacles, replicating typical scenarios where manipulators move around machinery, workpieces, and human workers.

Scenarios involving deep cavity insertion or complex, occluded shapes are excluded, as they exceed the sensing capabilities of fixed cameras and are beyond the scope of this work.

We consider various tool shapes attached to the \ac{EE} by defining a tool region and simulating point cloud within the region as illustrated in Fig.~\ref{fig:sim}B.

\subsection{Safety Critic}
We follow the approach by Srinivasan et al.~\cite{srinivasan2020learning}, which defines the discounted probability of future constraint violations. The optimal safety critic for the $i$-th constraint estimates the following expectation of the nominal controller:
\begin{equation}
V^{safe}_{C_i}(s_t) = c_i(s_t) + (1 - c_i(s_t)) \sum_{t'>t}^T \mathbb{E}\left[\gamma^{t'-t} c_i(s_{t'})\right].
\end{equation}

As specified in Table.~\ref{tab:rewards}, the cost function $c_i$ is an indicator function in our setup. The cumulative discounted probability of violations is estimated using Bellman equation:
\begin{equation}
	V^{safe}_{C_i}(s)  
= c_i(s) + (1 - c_i(s)) 
{\mathbb{E}}\left[
\gamma V^{safe}_{C_i}(s')
\right] \in [0, 1].
\end{equation}

We train it via value iteration during the encoder-decoder model training.

During the deployment, we aggregate safety critics to evaluate the overall collision risk. The maximum safety value between the robot body and the tool area is used as
\begin{equation}
V^{safe}_{C}(s_t) = \max(V^{safe}_{C_{body}}(s_t), V^{safe}_{C_{tool}}(s_t))
\end{equation}
where $V^{safe}_{C_{body}}$ represents the value of the robot's body, and $V^{safe}_{C_{tool}}$ represents that of the tool constraint (see~\ref{sec:constraint}).

\subsection{Constrained Policy Optimization} \label{def_cmdp}
\subsubsection{Background}
A control problem is typically modeled as a \ac{MDP} in \ac{RL} framework, which is defined by a tuple $(S, A, r, p, \mu)$. 
$S$ is the set of states, $A$ is the set of actions, $r: S \times A \times S \rightarrow \mathbb{R}$ is the reward function, $p: S \times A \times S \rightarrow [0,1]$ is the state transition probability and $\mu$ is the initial state distribution. 
The aim of \ac{RL} is to find a policy $\pi: S \mapsto \mathcal{P}(A)$ that maximizes
\begin{equation}
	\label{eq_return}
	J_R(\pi) = {\mathbb{E}}\left[\sum_{t=0}^{\infty} \gamma^t r(s_t,a_t,s_{t+1})\right],
\end{equation}
where $\gamma \in [0, 1)$ is the discount factor.
Here, $\mathbb{E}[\ldots]$ represents the average over a finite batch of samples. $s_0$ is sampled from an initial state distribution $\mu$ and trajectories are sampled using $\pi$. 

This is extended into \ac{CMDP} to handle physical constraints, augmented with a set $C$ of cost functions that capture constraint violations $\{c_1, \ldots, c_n\}$ and  corresponding limits $E = \{\epsilon_1, \epsilon_2, \ldots, \epsilon_n\}$~\cite{achiam2017constrained}.
Each $c_i: S \times A \times S \rightarrow \mathbb{R}$ maps ($s_t, a_t, s_{t+1}$) to the cost of the transition.
An optimal policy maximizes $J_R$, while keeping the discounted sum of future costs $c_i$ below their respective threshold $\epsilon_i$, yielding the constrained optimization problem:
\begin{equation}
	\label{eq_objectiveCMDP}
	\max_{\pi} \quad J_R(\pi) \quad \text{s.t.} \quad \forall i \in \{1,\ldots,n\}, \ J_{C_i}(\pi) \leq \epsilon_i,
\end{equation}
where $J_{C_i}(\pi) = \mathbb{E}\left[\sum_{t=0}^{\infty} \gamma^t c_i(s_t,a_t,s_{t+1})\right]$.

This framework has been shown to be effective for high-dimensional control problems in robotics in recent works~\cite{lee2023evaluation, kim2024not}.
Following the previous works, we employ the \ac{P3O} algorithm.

\subsubsection{CMDP for collision avoidance}

Here we define the action and observation spaces, as well as the reward and cost functions for our \ac{CMDP}. 
The state transition probabilities follow the rigid body dynamics of the robot.

The action space is defined by the residual changes to the robot's joint positions: $q_t^{des} = q_t + a_t$ where $a_t$ represents the action at time $t$ and $q_t$ represents current joint angles.
Action distributions are modeled using a Gaussian distribution.
The observation space includes the previous joint position history, the latent state from the voxel encoder, and the target \ac{EE} pose, represented in Cartesian coordinates and euler angles in the base frame.

The reward and cost functions are outlined in Table~\ref{tab:rewards}.
The reward function encourages \ac{EE} pose tracking while also regularizing the smoothness of the robot's movements. The cost functions are defined as indicator functions.

\begin{table}[t]
\centering
\begin{tabular}{l|l}
\hline
\multicolumn{2}{c}{Rewards} \\ 
\hline
1. Position tracking & $ 1.0 -  clip(\lvert {p_{EE}^{targ}} - p_{EE}\rvert ^2$, 0.0, 1.0)  \\
2. Orientation tracking & $1.0 - 2.0 * \arccos((q_{EE} \cdot (q_{EE}^{targ})^{-1})_w)$ \\
3. Command smoothness 1 &  $- 1.0 \times 10^{-3} \ \lvert\lvert q_{t}^{des} - q_{t-1}^{des} \rvert\rvert^2$ \\
4. Command smoothness 2 &  $- 1.0 \times 10^{-4}\ \lvert\lvert q_{t}^{des} - 2q_{t-1}^{des} + q_{t-2}^{des} \rvert\rvert^2$  \\
\hline
\multicolumn{2}{c}{Costs} \\ 
\hline
1. Collision & 1.0 if the robot body collides \\
\makecell[l]{2. Tool region violation \\  \quad (Protective mode only) } & 1.0 if any ESDF (corner point$_i$) $< 0.025$ \\ 
3. End-effector speed limit & 1.0 if EE speed exceeds 0.5 $m/s$ \\
4. Joint speed limit & 1.0 if $\rvert \dot{q}_{i,t} \rvert > 0.8  \cdot \dot{q}_{i, \text{lim.}}$ \\
\hline
\multicolumn{2}{c}{}\\ 
\end{tabular}
\caption{Reward and Cost Functions.}
\label{tab:rewards}
\end{table}

\subsubsection{Variable Tool Region Constraint}\label{sec:constraint}
The tool region, defined as a bounding box attached to the \ac{EE} (Fig.~\ref{fig:overview}A-ii), is randomized in each episode. We evaluate the \ac{ESDF} values at the eight corner points and flag a constraint violation if any value is less than half the voxel width.


\revise{
\subsection{Model Implementation}

\subsubsection{Architecture}
The architecture is shown in Fig.~\ref{fig:sim}. A voxelized point cloud is processed by three 3D \ac{CNN} layers and concatenated with proprioceptive inputs--including joint state history and user commands--before passing through a \ac{MLP} to produce a latent vector. Joint states help filter out the robot’s own body and assess collision risk.

\subsubsection{Scenario Setup}
All training data was generated in simulation using approximate robot and workspace geometry (Fig.~\ref{fig:sim}C), without relying on precise CAD models.

To define a new setup, users have to specify the workspace dimensions (x, y, z), grid resolution, sampling ranges for target \ac{EE} poses, and a range of tool region parameters, including size and offsets. Then the obstacles and tool bounding boxes are randomized each episode.

\subsubsection{Training}
We first collect data for training the encoder–decoder and safety critic by running a nominal controller (differential IK) in simulation.
Occlusion labels are assigned to grid cells where the \ac{ESDF} value is less than half the grid resolution.
These components are trained using supervised learning and value iteration.

The policy is then optimized using \ac{P3O}~\cite{shen2022penalized}, with the encoder–decoder and safety critic frozen to enable stable and efficient learning based on fixed perception and safety estimates in simulation.




}

\section{EXPERIMENTAL RESULTS}


We present real-world experiments and qualitative evaluations to validate each component of our system. See \href{https://youtu.be/lUzTVXjYM4k}{Supplementary video} for our main result.




\subsection{Experimental Setup}
The collision avoidance behavior and task-space position tracking performance of our system are evaluated using the setup illustrated in Fig.~\ref{fig:modes}A. We used the Indy7 robot, which is a collaborative robotic arm with six degrees of freedom and is equipped with a two-finger gripper.
The depth camera, an Intel RealSense D435, was mounted on the side of the table, as depicted in the figure. The wrist camera in the picture is not used for collision avoidance.

To assess the system's robustness in the presence of occlusions, we use a single point cloud source with a limited field of view (Realsense D435). This allows us to evaluate the system under realistic sensing constraints.

The robot operates within a workspace measuring 0.6 m ($x$-axis) by 1.0 m ($y$-axis) by 0.6 m ($z$-axis). Within this workspace, we construct a voxel grid with 0.05 m resolution.


    

    
\subsection{Use-case Example: Pick-and-Place}

Fig.~\ref{fig:intro} illustrates a pick-and-place system where our collision avoidance system is integrated.  The supplementary video demonstrates this behavior in action. We use a finite state machine to switch between picking and placing operations. 
For grasp planning, we used ANY-grasp~\cite{fang2023anygrasp}.
\revise{The Engage mode is activated when a grasp pose is detected by ANY-grasp. Protective mode is used otherwise.}
This setup demonstrates how our approach can be incorporated into existing pipelines.



\subsection{Computation Time}
Our system operates at \revise{50 Hz}.
The average update time for each action is approximately 6 ms, enabling the system to scale for deployment at frequencies exceeding 100 Hz.
All tests were performed on a desktop PC equipped with an RTX 3060 GPU (12 GB), an Intel i7-12700F CPU, and 32 GB of RAM.

Model inference averages 1.3 ms (SD 0.5 ms, max 4.0 ms), performing a single-pass evaluation of the encoder-decoder, policy, and safety critic. Other processes, including solving kinematics, communication, and observation computation, take 3.9 ms (SD 3.9 ms, max 19.0 ms). Point cloud processing—implemented with Warp~\cite{warp2022}—runs in parallel and averages 7.9 ms (SD: 3.7 ms, max: 72.3 ms) for voxelization, occupancy computation, and raycasting.

\begin{figure}
    \centering
    \includegraphics[width=0.95\columnwidth]{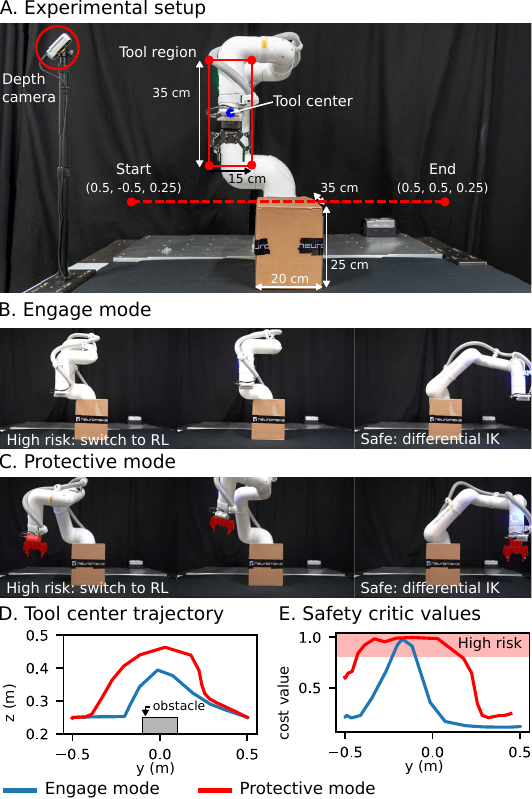} 
\caption{\textbf{Tool-Adaptive Collision Avoidance.} (A) Experimental setup. (B) Motion in Engage mode, where the tool area is permitted to make contact with the environment. (C) Motion in Protective mode, where the policy and the safety critic treat the tool region as part of the robot's body and avoid contact. (D) Tool center trajectories for each mode. (E) Corresponding safety critic values.
}
\label{fig:modes} 
\end{figure}

\subsection{Analyzing Adaptive Behavior}

A static obstacle was placed at the center of the workspace to analyze obstacle avoidance behavior, following a setup similar to \cite{khansari2012dynamical}. The robot was tasked with moving its \ac{EE} from the left to an endpoint on the right.

\subsubsection{Interaction Mode Behaviors}


Fig.~\ref{fig:modes}B–D illustrate the trajectories executed by the robot in each mode. In Protective mode, the robot accounts for the tool region, resulting in longer detours to avoid potential collisions.

\subsubsection{Safety Critic Values}
Fig.~\ref{fig:modes}E shows the cost values by the safety critic, calculated as the maximum of the body cost and the tool cost value.
The controller switches to the RL policy when the cost value exceeds a predefined threshold (0.8) and reverts to the nominal controller once the cost decreases.



\subsubsection{Estimated Collidable Region}
\begin{figure}
    \centering
    \includegraphics[width=\columnwidth]{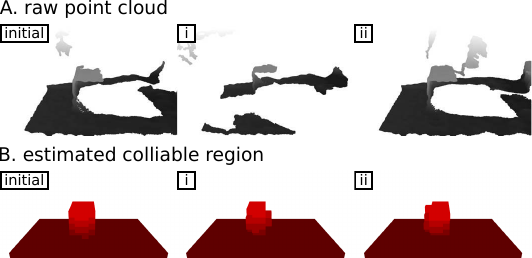} 
\caption{\textbf{Visualization of Point Cloud and Collidable Region Estimation.}
Front view snapshots at the initial configuration, at the start (i) and at a mid (ii) point of the trajectory.
(A) Raw point cloud data captured during the experiment.
(B) Corresponding estimated collidable regions.}
    \label{fig:voxel} 
\end{figure}

Fig.~\ref{fig:voxel} shows the raw point cloud and the estimated occupancy during the experiment in Fig.~\ref{fig:modes}. 
When the gripper is at the left side, the table is significantly occluded (Fig.~\ref{fig:voxel}A,B-i). Our model remembers the measurements at the initial configuration and preserves the occupancy prediction.

Our model filters out measurements corresponding to the robot's body and tool components, such as the gripper, camera, and cables. This capability is achieved by incorporating randomly simulated tool geometry during training.

\subsubsection{Adaptation to Tool Shape}
Fig.~\ref{fig:errors}A-C depict the task-space position tracking performance in the $yz$ plane for three different \ac{EE} configurations.
Each target position was maintained for three seconds \revise{after convergence}, and the average tracking error was computed. The experiment follows the setup shown in Fig.~\ref{fig:modes}A, with errors measured via forward kinematics. 

As collision avoidance is activated, higher tracking errors are observed near the table and obstacle. In contrast, open areas consistently exhibit errors around 0.01 mm. 
As larger tool areas were commanded, the high-error region expands \revise{(Fig.~\ref{fig:errors}B and Fig.~\ref{fig:errors}C)}. This demonstrates the system's ability to adaptively predict potential collisions based on tool dimensions and behave accordingly.

\subsection{Validation of the Control Strategy}

\begin{figure}
    \centering
    \includegraphics[width=0.95\columnwidth]
    {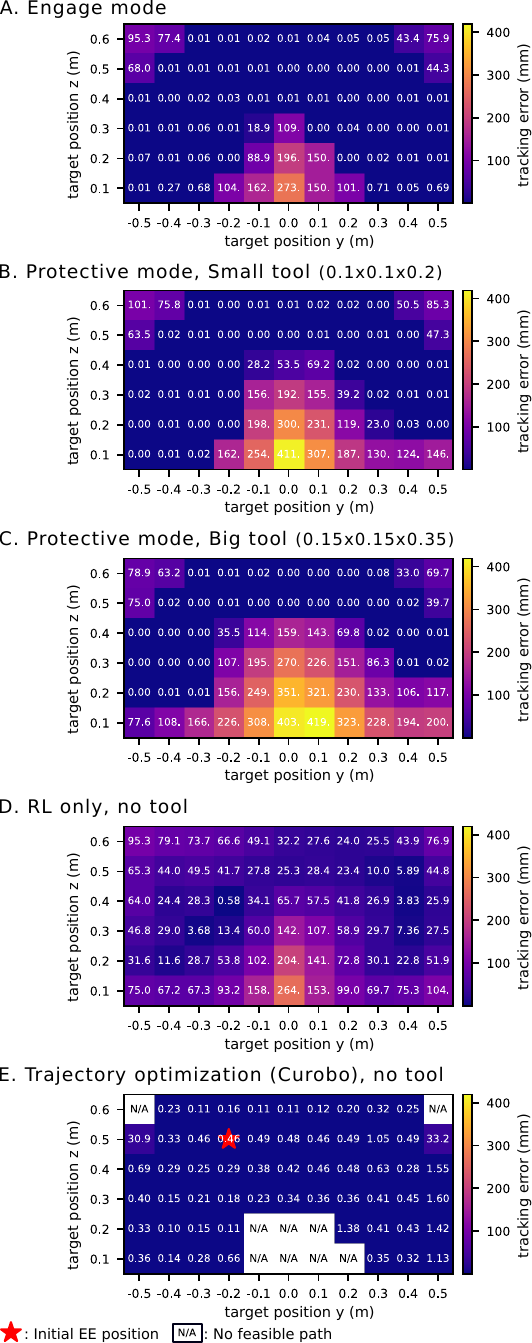}
    \caption{\textbf{Tracking Errors Across Different Tool Conditions and Control Approaches}. Task-space position tracking error in the $yz$ plane, with target positions in a 0.1 m grid at a fixed $x$-coordinate of 0.5 m. 
(A) Engage mode: only robot body collisions are considered.
(B) Protective mode: results for different tool sizes.
(C) RL-only approach.
(D) Trajectory optimization.
    } 
    \label{fig:errors} 
\end{figure}

Fig.~\ref{fig:errors}D illustrates the performance of RL alone, where tracking errors in the safe area are significantly higher compared to ours (Fig.~\ref{fig:errors}A).
This highlights the inherent limitations of RL in achieving the precision offered by traditional \ac{IK} methods.

We also compared our system to Curobo~\cite{sundaralingam2023curobo}, a state-of-the-art GPU-accelerated \revise{\ac{TO}} framework (Fig.~\ref{fig:errors}E).
The results are similar: Curobo also reports no feasible solution (marked as N/A) in obstacle regions.
\revise{In collision-free regions, our method achieves lower tracking error because of the refinement using iterative IK}.

\begin{figure}
    \centering
    \includegraphics[width=0.95\columnwidth]{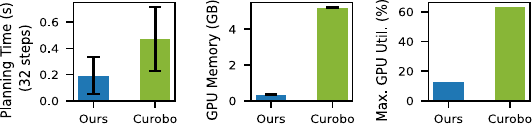} 
\caption{\textbf{Resource Usage Compared to Curobo (with NvBlox)~\cite{sundaralingam2023curobo}}}
    \label{fig:curobo} 
\end{figure}

Our system requires fewer computational resources overall, as shown in Fig.~\ref{fig:curobo}.
The average planning time \revise{for Curobo} is 468 ms (SD 244 ms, max 1.36 s)\revise{, whereas our method operates within 10 ms. In terms of memory usage, ours requires only 0.45 GB of GPU memory, compared to approximately 4.5 GB for Curobo when running with NvBlox.}
It is important to note that \revise{this} does not include voxel grid mapping and other system components. 

While global optimizers~\cite{sundaralingam2023curobo,fishman2024avoid} excel at long-horizon planning under full observability, our lightweight system is better suited for reactive, short-horizon collision avoidance—especially when computational resources are shared with high-level processes.


\begin{figure}
    \centering
    \includegraphics[width=0.98\columnwidth]{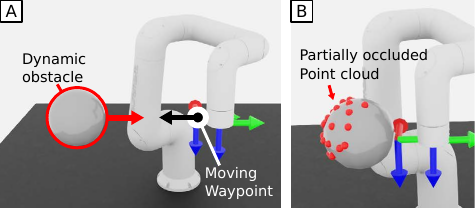} 
\caption{\textbf{Dynamic obstacle avoidance experiment} (A) Experimental setup. The obstacle and waypoint move in opposite direction. The waypoint moves at \unit[0.25]{m/s}. (B) Visualization of point cloud on the obstacle}
    \label{fig:dyn} 
\end{figure}


\subsection{Evaluation in Dynamic Settings}
\label{sec:dynamic}
To assess the reliability of our system, we evaluate its performance in dynamic environments with moving obstacles and compare it against widely used baseline methods.

\subsubsection{Experimental Setup}
As shown in Fig.~\ref{fig:dyn}A, the \ac{EE} is commanded to move from right to left at a speed of 0.25 m/s. For the dynamic case, an obstacle is spawned on the left side of the workspace and moves at a fixed linear velocity toward the center of the \ac{EE} target trajectory.
In the static case, the obstacle is fixed at the center. Each method is evaluated over 50 trials.

All controllers operate at 50 Hz; to ensure consistent timing, the simulation pauses each step until computation is complete.

\subsubsection{Baselines}
We validated our control strategy against two baseline approaches with different perception methods:

\begin{itemize} \item \textbf{\ac{APF}}: A classical reactive path planning method~\cite{khatib1986real}.
\item \textbf{\ac{MPPI}}: A sampling-based \ac{MPC} method~\cite{sundaralingam2023curobo,liu2022regularized, vasilopoulos2023ramp}. We followed STORM~\cite{storm2021} implementation with a 0.5 s horizon and 4000 samples.
\end{itemize}

Traditional approaches often preprocess point clouds by clustering and approximating obstacles as ellipsoids or convex hulls~\cite{becker2025informed, de2017multimodal}. More recent work has adopted learned \ac{ESDF} or collision check for improved robustness and reactivity~\cite{liu2022regularized, danielczuk2021object}, which aligns with our learning-based approach.

For evaluation, we implemented two perception variants for the baselines:
(i) point cloud obstacle approximation using convex hulls~\cite{de2017multimodal, rakita2021collisionik} combined with Kalman filter-based centroid tracking, and 
(ii) a learned SDF model~\cite{liu2022regularized}. 

\subsubsection{Result}
Table~\ref{tab:baselines} summarizes the performance of our method compared to baseline approaches under varying obstacle speeds. \textbf{GT} denotes access to ground-truth obstacle position and shape.

Our method achieves performance comparable to APF with learned SDF~\cite{liu2022regularized}. Across all scenarios, approaches using learned perception demonstrate higher robustness to dynamic changes and partial observability.

APF occasionally fails in static settings due to local minima and may exhibit oscillatory behavior caused by sudden, high repulsive forces.
MPPI shows strong performance in static scenarios but struggles with dynamic obstacles. Despite running at high frequency without delay, its trajectory samples requires time to converge to collision-free trajectories, which limits responsiveness.

The convex hull approximation approach suffers significantly from partial occlusions, as illustrated in Fig.~\ref{fig:dyn}B. In such cases, the obstacle is underestimated in size, leading to increased collision rates.

\subsubsection{Scalability}
 Traditional methods involving point cloud clustering and filtering~\cite{becker2025informed, de2017multimodal} have a per-cycle runtime that grows with the number of segmented obstacles. By contrast, our method and learned approaches using the whole scene~\cite{liu2022regularized, danielczuk2021object} maintain a constant runtime.

\setlength{\tabcolsep}{3pt}  
\begin{table}[t]
    \centering

    \caption{\textbf{Collision rate (Coll.) and success rate (Succ.) for different methods under static and dynamic obstacle conditions.} All values in \%}
\begin{tabular}{ll|cc|c|c}
\hline
\multicolumn{2}{c|}{Method} & \multicolumn{2}{c|}{Static} & 0.2 m/s & 0.4 m/s \\
\cline{3-6}
Perception & Control        & Succ. & Coll.              & Coll.  &Coll.  \\
\hline
\multirow{2}{*}{GT Pose}     
  & APF                 & 100  & 0 & 2  & 18 \\
  & MPPI                & 100  &  0 & 6 & 56 \\
  
\hline
\multirow{2}{*}{Convex hull~\cite{de2017multimodal, rakita2021collisionik}} 
  & APF                 & 54  & 16 & 78 & 82 \\
  & MPPI                & 72  &  0 & 50 & 72 \\
  
\hline
\multirow{2}{*}{Learned 
SDF~\cite{liu2022regularized}} 
  & APF                 & 96  & 4 & 8 & 38 \\
  & MPPI                & 100  &  0 & 16 & 65 \\
  
\hline
\multirow{1}{*}{Ours (Occupancy grid) } 
  & RL               & \textbf{100} &  \textbf{0} & \textbf{6}  & \textbf{36} \\
\hline
\end{tabular}

    \label{tab:baselines}
\end{table}

\subsection{Effectiveness of the Learned Perception Model \revise{for RL}}

\begin{table}[t]
    \centering
    \caption{\textbf{Comparison across different scene representations. } \\All values in \%}
    \begin{tabular}{l|ccc|cc}
    \hline     
    \multicolumn{1}{c|}{Method}  & \multicolumn{3}{c|}{Protective Mode} & \multicolumn{2}{c}{Engage Mode} \\ 
         \cline{2-6}
         & Succ. & Coll. & Tool Coll. & Succ. & Coll. \\ 
        \hline
        End‑to‑End       & 10.0  & 2.51 & 5.03 & 11.0  & 6.48 \\ 
        Autoencoder (AE) & 47.7  & 0.86 & \textbf{1.27} & 59.3  & 1.68 \\ 
        Mapping + AE     & 54.1  & 1.06 & 2.26 & 72.8  & 2.44 \\ 
        Ours             & \textbf{67.3} & \textbf{0.42} & 1.31 & \textbf{82.5} & \textbf{0.82} \\ 
        \hline
    \end{tabular}

    \label{tab:merged_col_rate}
\end{table}

We evaluated our perception approach, which uses an encoder-decoder model to reconstruct collision regions, filter out noise, and remove robot and tool areas from the voxel grid. 

\subsubsection{Baselines}  
We compared three methods:

\begin{itemize}
    \item \textbf{End-to-End Training:} Training directly from scratch using the same RL algorithm without pretraining.
    \item \textbf{Pretrained Autoencoder (AE):} A 3D CNN is pretrained to reconstruct the input voxel grid, and its latent representation is used as input to the policy.
    \item \textbf{Mapping with Autoencoder (Mapping+AE):} 
    This approach explicitly filters out robot and tool points from the raw point cloud, constructs a voxel-based occupancy grid, and uses it as input to the policy network. We integrated Voxblox~\cite{oleynikova2017voxblox} into the simulation, allowing the policy to use the voxel map as external memory.
\end{itemize}

\subsubsection{Results}  

The results, summarized in Table~\ref{tab:merged_col_rate}, are evaluated in both control modes in simulation. In Protective mode, tool shape is randomized. The tool violation rate reflects the percentage of episodes where tool constraints were violated. Metrics are computed from 20,000 episodes, with waypoints sampled from free space (\ac{ESDF} $>$ 0.1) in the workspace. Each episode runs for 10 seconds.

End-to-end training with \ac{P3O} yields suboptimal performance, resulting in overly conservative behavior that reduces success rates and increases collision frequency.

While the pretrained encoder (AE) improves over end-to-end training, it still leads to lower success rates and higher collision rates compared to our method.

Although the Mapping+AE approach achieves a high success rate, it struggles with perception uncertainties such as noise and occlusions. A key failure mode we observed is that robot and tool filtering occasionally removes nearby obstacle points when they are too close to the robot, resulting in unexpected collisions.

\section{CONCLUSION}
We presented a fast and tool-aware collision avoidance system for collaborative robots.
A use case example, shown in Fig.~\ref{fig:intro}, highlights its applicability to realistic scenarios involving collaborative robots under practical sensing constraints.
By integrating constrained \ac{RL} with a traditional differential IK approach, our system achieves both robustness and high accuracy. 
It seamlessly adapts to diverse tool shapes and interaction modes, ensuring safe and reliable operation while \revise{supporting} real-time performance \revise{exceeding 100 Hz}

\revise{Compared to baseline methods such as APF and MPPI, our system consistently achieves lower collision rates and comparable or higher success rates in dynamic environments. Experimental results further show robust performance with control latency under 10 ms and GPU memory usage of only 0.45 GB—over 10 times more memory-efficient and significantly faster than  Curobo's \ac{TO}~\cite{sundaralingam2023curobo}.}

Future work will focus on integrating the system into contact-rich and complex manipulation frameworks such as ALOHA~\cite{zhao2024aloha} to study high-level intelligence. Additionally, we aim to extend the system to mobile manipulators and multi-robot scenarios operating in variable workspaces.

\section*{ACKNOWLEDGMENT}
We appreciate Dayoung Kim for helping camera setup and calibration.


\bibliographystyle{bibliography/IEEEtran}

\end{document}